\documentclass[journal]{IEEEtran} % use the `journal` option for ITherm conference style
\IEEEoverridecommandlockouts
\usepackage{cite}
\usepackage{amsmath}
\usepackage{graphicx} % Required for inserting images

\title{\textbf{SEMANTIC NEURAL MODEL APPROACH FOR FACE RECOGNITION FROM SKETCH}}
\author{CHANDANA NAVULURI, SANDHYA JUKANTI, RAGHUPATHI REDDY ALLAPURAM}

\begin{document}
\maketitle

\textbf{ABSTRACT}:

\textit{Face sketch synthesis and reputation have wide range of packages in law enforcement. Despite the amazing progresses had been made in faces cartoon and reputation, maximum current researches regard them as  separate responsibilities. On this paper, we propose a semantic neural version approach so that you can address face caricature synthesis and recognition concurrently. We anticipate that faces to be studied are in a frontal pose, with regular lighting and neutral expression, and have no occlusions. To synthesize caricature/image photos, the face vicinity is divided into overlapping patches for gaining knowledge of. The size of the patches decides the scale of local face systems to be found out.}


\textbf{\textbf{I. INTRODUCTION}}

 Face recognition is a way for spotting or confirming an character's identity through looking at their face. Face popularity software program can recognize people in snap shots, videos, or in real time. Due to its critical applications in regulation enforcement, automatic face sketch-to-picture reputation has continually been a high precedence in computer imaginative and prescient and device getting to know. During police stops, officers can also use cell devices to pick out individuals. In numerous crook and intelligence investigations, a forensic hand-Drawn or laptop-generated composite caricature based totally at the description given by way of eyewitness testimony is the most effective clue to identify capacity   Suspects. As a result, an automated matching algorithm is needed to experiment law enforcement face databases or surveillance cameras the usage of a forensic comic strip quick and accurately. The forensic or composite drawings, then again, most effective encompass some simple details about the suspects' appearance, inclusive of the spatial topology in their faces, while other smooth biometric traits, consisting of pores and skin colour, race, or hair color, are omitted. Traditional sketch recognition algorithms are divided into two kinds: generative and discriminative algorithms. Till matching, generative techniques switch one of the modalities into the other. Discriminative strategies, such as the dimensions-invariant function remodel, on the other hand, extract features (sift). These characteristics, however, are not usually ideal for a cross-modal reputation venture. As a result, different techniques for gaining knowledge of or extracting modality-invariant features are being investigated. Deep mastering-based totally approaches, which analyze a famous latent embedding among the 2 domains, have lately emerged as potentially possible strategies for tackling the go-domain face popularity issue. Deep mastering techniques for caricature-to-photo reputation, on the other hand, are extra tough to apply than for other single-modality domains due to the fact they require a huge number of facts samples to prevent overfitting and nearby minima. Moreover, there are just a few hundred caricature-image pairs in the modern publicly handy comic strip-image datasets. Extra importantly, maximum datasets simplest have one cartoon according to topic, making it hard, if now not not possible, for the community to examine sturdy latent functions. As a result, maximum techniques have used a community that is both too shallow or simplest skilled on one of the modalities (typically the face photo). Current brand new tactics are especially involved with remaining the semantic representation gap among the 2 domains at the same time as ignoring the dearth of smooth-biometric knowledge in the comic strip modality. Despite the brilliant effects of latest sketch-photograph recognition algorithms, there may be still a step lacking on this segment: conditioning the matching method on tender biometric traits. There are usually a few facial attributes lacking within the sketch area, which includes skin, hair, and eye shades, gender, and ethnicity, in particular inside the software of comic strip-photo popularity, that is based on the quality of sketches. In precis, the main contributions of this paper consist of the following:
• we suggest a semantic neural version method for face reputation from comic strip.
• to improve the efficiency of our comic strip photo popularity, we enforce a new loss characteristic that fuses the facial attributes given through eyewitnesses with the geometrical residences of forensic sketches.
• we use switch gaining knowledge of algorithm to extract the capabilities and use these functions to extract other capabilities to get correct consequences.

\textbf{\textbf{II. LITERATURE REVIEW}}

\textbf{~\cite{kazemi2018attribute}}:
Multi-Scale Gradients Self-Attention Residual Learning for Face          Photo-Sketch Transformation
Face sketch recognition, has made considerable progress in recent years. Due to the difference of modality between face photo and face sketch, traditional exemplar-based methods often lead to missed texture details and deformation while synthesizing sketches. And limited to the local receptive field, Convolutional Neural Networks-based methods cannot deal with the interdependence between features well, which makes the constraint of facial features insufficient; as such, it cannot retain some details in the synthetic image.
Moreover, the deeper the network layer is, the more obvious the problems of gradient disappearance and explosion will be, which will lead to instability in the training process. Therefore, in this paper, we propose a multi-scale gradients self- attention residual learning framework for face photo- sketch transformation that embeds a self-attention mechanism in the residual block, making full use of the relationship between features to selectively enhance the characteristics of specific information through self- attention distribution. Simultaneously, residual learning can keep the characteristics of the original features from being destroyed.
In addition, the problem of instability in GAN training is alleviated by allowing discriminator to become a function of multi-scale out- puts of the generator in the training process. Based on cycle framework, the matching between the target domain image and the source domain image can be constrained while the mapping relationship between the two domains is established so that the tasks of face photo-to-sketch synthesis (FP2S) and face sketch- to- photo synthesis (FS2P) can be achieved simultaneously. Both Image Quality Assessment (IQA) and experiments related to face recognition show that our method can achieve state-of- the-art performance on the public benchmarks, whether using FP2S or FS2P.

{Some of the drawbacks of this model } 

•	Difficulties to obtain better performance 

•	Inaccurate estimations of the missing pixels

•	High prediction complexity for large datasets 

•	Higher prediction complexity with higher dimensions 

\textbf{~\cite{galea2017forensic}}:Graph-Regularizd Locality-Constraine Join  Dictionary and Residual Learning for Face Sketch Synthesis[Junjun Jiang, Yi Yu, Zheng Wang, Xianming Liu 2019]
Most of the current face sketch synthesis approaches directly learn the relationship between the photos and sketches, and it is very difficult for them to generate the individual specific features, which we call rare characteristics

{DRAWBACKS:}
Cannot improve accuracy by preserving fast processing.

Cannot achieve noise resistant detection.

Classification accuracy is lower

\textbf{~\cite{zhao2018deep}: }Cross-Domain Face Sketch Synthesis[MINGJIN ZHANG, JING ZHANG, YUAN CHI, YUNSONG LI 2019]
In the proposed cross-do domain, while the target task is to recover the structure in the sketch domain. But in reality, the training data is not suficient to learn the model main synthesis work, the source task is to construct the structure of faces in the photo

{DRAWBACKS:}Method is sensitive to noise.

Cannot improve accuracy by preserving fast processing.

Cannot achieve noise resistant detection.

\textbf{~\cite{4}:}A Deep Collaborative Framework for Face Photo–Sketch Synthesis[Mingrui Zhu, Jie Li, Nannan Wang 2019]
This strategy can constrain the two opposite mappings and make them more symmetrical, thus making the network more suitable for the photo–sketch synthesis task and obtaining higher quality generated images. Qualitative and quantitative experiments demonstrated the superior performance of our model in comparison with the existing state-of-theart solutions.

{DRAWBACKS}

Solutions have been proved ineffective.

Imbalance classification is the most critical and a well-known problem.

Extremely difficult for the classification algorithm to predict.

\textbf{III. PROPOSED METHODOLOGY}

\textbf{A. SYSTEM OPERATION}

This paper aims at designing a framework that is able to simultaneously synthesize realistic face sketch image with which the domain discrepancy is reduced and extract discriminative face feature for face sketch recognition. The overall architecture mainly consists of two parts: namely a generator and a discriminator. The generator is fed with a face photo image and a corresponding sketch image can be obtained. For the discriminator, in order to learn the ability of face feature extraction, three images compose a triplet sample as the input (a generated fake sketch image, a ground-truth sketch image, and a hard-negative sketch image which has small distance with the ground-truth sketch).
Connected to the basic discriminator network, two branches are designed to implement the functions of real/fake sketch discrimination and face feature extraction. The discrimination branch is a convolutional layer with output size of 7×7 and a sigmoid activation layer to predict probability scores between 0 and 1, which is utilized to distinguish the input sketch image is true or fake. The face feature extraction branch is a fully-connected layer with output size of 1024, with which a face sketch image can be represented as a 1024-dimension feature vector.
The input nodes receive the input in the form of numeric expression. The information is represented as activation values and passed through the hidden layers to reach the output nodes. A Feed-Forward Network topology is implemented were, the signal travels in only one direction. Each element process computation based on weighted sum of the input and compares it to a nodes and newly calculated value is feed to the next layer. The mathematical representation is expressed . For sketch-based face classification a Multilayer Feed Forward Network has been adopted. In this type network is made up of one or more hidden layers, whose computation nodes are called hidden neurons or hidden units. We use python language to code this model and have imported keras as a deep learning component to extract and train the system.

\includegraphics[width=0.5\textwidth]{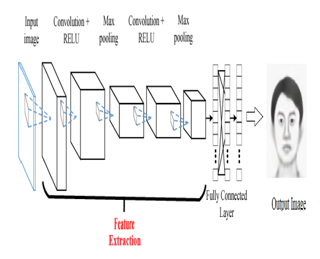}

\textbf\textbf{{FIGURE 1. FACE SKETCH RECOGNITION MODEL}
}

 \textbf{\textbf{B. MODULES}}

There are two main modules in this model:

i.	Pre-processing

ii.	Feature Extraction

\textbf{i.PRE-PROCESSING}
Pre-processing simply means that after one algorithm has been applied to the image, the output of that algorithm is fed into the input of other algorithms, with the end result being an improved image. Of course, a pre-processing phase can be used in an image enhancement algorithm, and an improved image can still be used as an input for other algorithms.
Convert colour images to grayscale to minimise computation complexity: in some cases, losing redundant details from your images to save space or reduce computational complexity is a good idea.Converting coloured images to grayscale images, for example. This is because colour isn't always needed to identify and perceive an image in many objects. Grayscale may be sufficient for identifying such artefacts. Colour images, which contain more details than black and white images, can add needless complexity and take up more memory space (Keep in mind that colour images are expressed in three channels, so converting them to grayscale decreases the number of pixels that must be processed.).
The need to resize the images in your dataset to a unified dimension is a significant restriction in some machine learning algorithms, such as CNN. This means that before feeding our images to the learning algorithm, they must be pre-processed and scaled to have equal widths and heights.
Another common pre-processing technique is to add perturbed versions of existing images to the existing dataset. Typical affine transformations include scaling, rotations, and other affine transformations. This is done to expand your dataset and expose your neural network to a wide range of image variations. This increases the likelihood that your model will identify objects in some shape or type.

\includegraphics[width=0.5\textwidth]{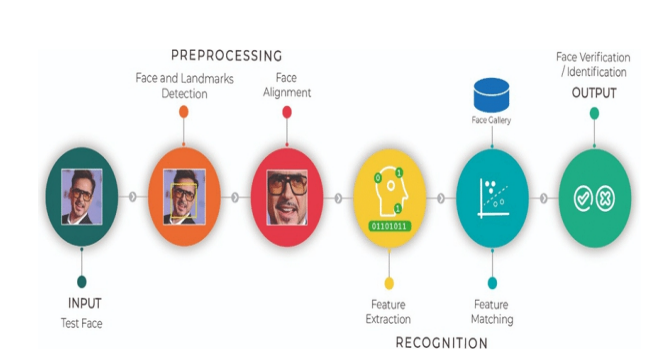}

\textbf\textbf{{\textbf{FIGURE 2.FLOW CHART OF MODULE1}}
}

\textbf{\textbf{ii. FEATURE EXTRACTION}}

Feature extraction plays a few transformation of original features to generate different functions which can be extra significant. 
Feature extraction can be used in this context to reduce complexity and supply a easy illustration of facts representing each variable in characteristic area as a linear combination of authentic input variable. The most famous and widely used feature extraction approach is precept component evaluation.
Records advantage (ig) is a metric that compares the increase in entropy when a feature is gift to whilst it isn't. That is the utility of greater general techniques, such as informational entropy calculation, to the trouble of figuring out the importance of a characteristic in characteristic area.Characteristic that is based on correlation choice looks for function subsets based at the diploma of characteristic redundancy. The goal of the assessment is to discover subsets of functions which can be fairly correlated with the class personally but have low inter-correlation. The significance of a network of capabilities will increase as the correlation between capabilities and class rises, and reduces because the inter-correlation grows. Cfs is often used along side search methods consisting of forward choice, backward exclusion, bi-directional search, quality first search, and genetic search to decide the first-class feature subset.Clear out’s method makes use of diverse scoring functionalities and pick out pinnacle-n functions having the best ratings. They are computationally quicker than wrapper method. The trouble is that the characteristic dependencies aren't considered

\includegraphics[width=0.5\textwidth]{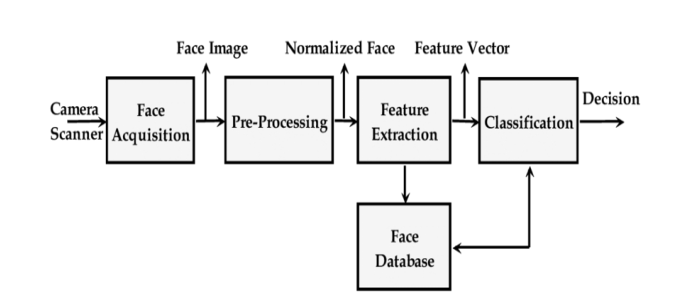}

\textbf\textbf{{\textbf{{10pt}FIGURE 3.FLOW CHART OF MODULE2}}
}

\textbf{convolutional layer in a CNN can be expressed as follows}
\begin{equation*}
    \mathbf{h}^{(l)}_{i,j,k} = g \left(\sum_{m=1}^{M^{(l-1)}}\sum_{p=1}^{P^{(l)}}\sum_{q=1}^{Q^{(l)}}\mathbf{W}^{(l)}_{p,q,m,k} \cdot \mathbf{h}^{(l-1)}_{(i-1+s^{(l)}p),(j-1+s^{(l)}q),m} + \mathbf{b}^{(l)}_{k} \right)
\end{equation*}

where:
the activation of the $k$th filter in the $l$th layer at position $(i,j)$

- $\mathbf{h}^{(l)}_{i,j,k}$

- $g(\cdot)$ is the activation function

the weight between the $m$th channel in the $(l-1)$th layer and the $k$th filter in the $l$th layer at position $(p,q)$
- $\mathbf{W}^{(l)}_{p,q,m,k}$ 

 the activation of the $m$th channel in the $(l-1)$th layer at position $(i-1+s^{(l)}p,j-1+s^{(l)}q)$
 
- $\mathbf{h}^{(l-1)}_{(i-1+s^{(l)}p),(j-1+s^{(l)}q),m}$  

- $\mathbf{b}^{(l)}_{k}$ is the bias term for the $k$th filter in the $l$th layer

- $M^{(l-1)}$ is the number of channels in the $(l-1)$th layer
- $P^{(l)}$ and $Q^{(l)}$ are the height and width of the filters in the $l$th layer, respectively
- $s^{(l)}$ is the stride of the filters in the $l$th layer

\textbf{C. ALGORITHM}

For this model we use Transfer Learning Algorithm. This algorithm has advantages like less training data, models generalize better and makes deep learning more accessible. The reuse of a pre-trained model on a new problem is known as transfer learning in machine learning. A computer uses the information learned from a previous task to enhance generalization about a new task in transfer learning. We train this model using this algorithm and dataset of photos and corresponding sketches. This algorithm extracts a feature and starts to extract a new feature with already extracted feature. For example, if the algorithm extracted the eye part of the face, it starts to extract another feature like nose or lips with matching eyes which is already extracted. This algorithm helps in extracting the exact photo using the feature extraction.
\includegraphics[width=0.5\textwidth]{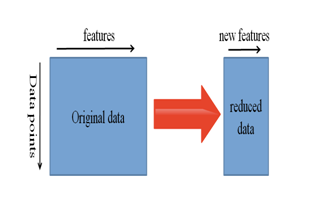}
\textbf{{\textbf{FIGURE 4. TRANSFER LEARNING ALGORITHM}}
}

\textbf{Accuracy and Loss functions}

\textbf{Accuracy} = (Number of Correct Predictions / Total Number of Predictions) x 100%
 
\textbf{Mean Squared Error (MSE) } 

MSE = $\frac{1}{n} \sum_{i=1}^{n} (y_i - \hat{y_i})^2$

where N is the number of samples in the dataset, y is the true label (0 or 1), and yhat is the predicted probability for that sample.

\textbf{IV BACKGROUND}

i. SEMANTIC NEURAL MODEL:
Semantic neural model approach refers to a class of machine learning models that are designed to learn and represent the meaning or semantics of input data, such as text, images, or other forms of structured or unstructured data. 
semantic neural model is the convolutional neural network (CNN)  
CNNs use a series of convolutional layers to extract features from images , which can then be used to identify objects, recognize faces, or perform other tasks.

ii.IMAGE RECOGNITION FROM SKETCH:
When it comes to recognizing faces from sketches, there are several approaches that can be taken using semantic neural models. 
We use a convolutional neural network (CNN) to extract features from both the sketch and the image of the person being identified.

\textbf{iii.SOFTWARE REQUIREMENT:}

We use Anaconda is a free and open-source delivery of the Python and R programming languages for use in data science and machine learning, with the goal of simplifying package control and deployment. Conda, the package control system, is used to handle package variants.

\textbf{\textbf{V. EXPERIMENTS AND ANALYSIS}}

we aimed to tackle the challenging tasks of face photo-sketch recognition by considering them as a face photo-sketch transformation problem. To address the instability in the training process based on GAN, We embedded a self-attention unit in the residual block to enable the generator to focus on facial features while reducing the interference of the background area, resulting in a generated image with richer texture information.
 The experimental results showed that our approach achieved significant improvement in terms of image quality assessment (IQA) and recognition accuracy of reconstructed images.As can be seen in Fig. 5, in the recognition experiment
i.implementation
We proposed the neural network architecture which is an encoder-decoder type of model. The encoder is a convolutional neural network (CNN) that takes the input grayscale image and converts it into a lower-dimensional representation. The decoder is also a CNN that takes the encoded representation and produces the output image.

The encoder network consists of several convolutional layers with different numbers of filters, kernel sizes, strides, and activation functions. The output of each convolutional layer is passed through a max-pooling layer to reduce the spatial dimensions. The decoder network is similar to the encoder but in reverse order, where each convolutional layer is followed by an upsampling layer to increase the spatial dimensions.

The loss function used is the mean squared error (MSE) between the predicted output image and the ground truth image. The optimizer used is the Adam optimizer, which is a popular optimization algorithm for training deep learning models.

\includegraphics[width=0.5\textwidth]{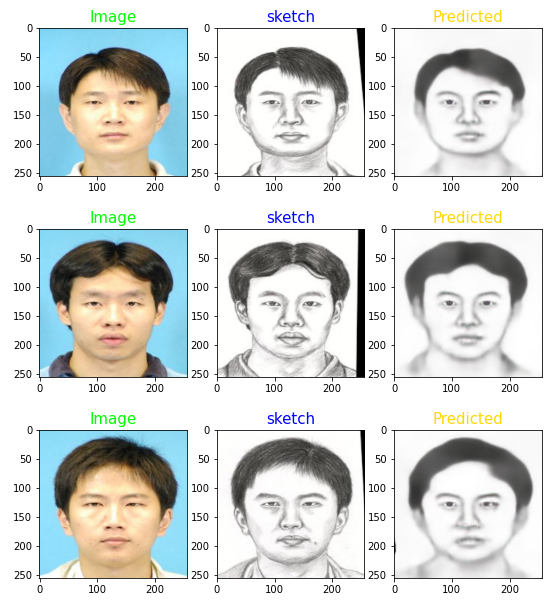}
\textbf{Figure: 5 face photo-sketches synthesized by our proposed
method }

{}

\textbf{VI.PERFORMANCE METRICS}

Face sketch synthesis, as a key technique for solving face sketch recognition, has made considerable progress in recent years.Due to the difference of modality between face photo and face sketch, traditional exemplar-based methods often lead to missed texture details and deformation while synthesizing sketches.When comparing with real-time photo the characteristics of the face are unrecognizable.Therefore, the objective is to find the original photo with a given sketch with maximum accuracy.

% \includegraphics[width=0.5\textwidth]{graph.png}
% \textbf{Figure: 6 Analysis of accuracy for different epochs}

\includegraphics[width=0.5\textwidth]{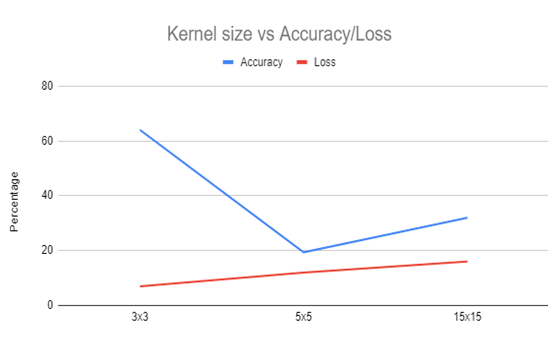}
{  }\textbf{Figure: 7  Accuracy for different kernel sizes}

\textbf{VIII. Guideline for Production Grade Deployment}

The "Semantic Neural Model Approach for Face Recognition from Sketch" project can be deployed on a cloud environment when there is a need for large scale usage. This would be especially relevant if the project needs to process a large number of sketches and images, or if it needs to handle a high number of concurrent requests. Deploying the project on the cloud also offers several advantages, such as scalability, reliability, and cost-effectiveness.

An example use case for this project could be a law enforcement agency that needs to identify suspects from hand-drawn sketches. They might receive thousands of sketches each day from different sources, such as eyewitnesses, security cameras, or composite sketch artists. The sketches would need to be quickly and accurately matched to a database of known individuals in order to help solve crimes.

To deploy the "Semantic Neural Model Approach for Face Recognition from Sketch" project on a cloud environment, the following steps could be taken:

Choose a cloud provider: There are several cloud providers available, such as Amazon Web Services (AWS), Google Cloud Platform (GCP), and Microsoft Azure. Choose a provider based on your specific requirements, such as cost, location, and features.

Select a deployment model: Decide whether to use a Platform as a Service (PaaS) or Infrastructure as a Service (IaaS) model. A PaaS model provides a pre-configured environment for deploying the project, while an IaaS model offers more control and flexibility over the deployment environment.

Set up the deployment environment: Set up a virtual machine (VM) or container to host the project. Install any necessary software and dependencies, such as Python, TensorFlow, and OpenCV.

Deploy the project: Copy the project files to the deployment environment and run the necessary scripts to start the project.

Set up scaling and monitoring: Configure the deployment environment to automatically scale based on usage patterns, and set up monitoring tools to track performance and detect issues.

Once the project is deployed on the cloud environment, the law enforcement agency can send sketches to the system for processing. The system would use the trained neural network to extract features from the sketches and match them to a database of known individuals. The agency could access the results through a web interface or API.

\textbf{Infrastrcuture Requirement}

The infrastructure requirements for deploying the "Semantic Neural Model Approach for Face Recognition from Sketch" 
project on a cloud environment would depend on factors such as the size of the dataset, the complexity of the neural 
network, and the expected usage patterns. However, in general, the following infrastructure components would be required:

CPU: The project would require a powerful CPU to process the sketches and images. The CPU would need to have 
multiple cores and a high clock speed to handle the computations efficiently.

GPU: A GPU would be beneficial for speeding up the training and inference of the neural network. 
The GPU would need to have a high number of CUDA cores and a large amount of memory to handle the large matrix operations involved in deep learning.

Storage: The project would require a large amount of storage to store the sketches, images, 
and neural network models. The storage should be fast and reliable to minimize data access times and prevent data loss.

Network QoS: A high-quality network with low latency and high bandwidth would be required to 
handle the large data transfers between the client and the cloud infrastructure. 
Quality of Service (QoS)~\cite{cherkasova2002measuring} mechanisms can be used to prioritize the traffic 
and ensure that critical data is delivered quickly. This is not only important for data movement but also 
necessary for fast and in-time response from the sketch identification system~\cite{liu2023intelligent, robin2021p4kp}. 

Load balancer: A load balancer~\cite{miao2017silkroad,nguyen2020toward,robin2022clb} would be necessary to distribute the incoming requests across
 multiple instances of the project to ensure high availability and scalability. The load balancer 
 would need to be highly available and capable of handling large amounts of traffic.

\textbf{VII. CONCLUSION}

The experiment was successful and the model returned the image as a grayscale image with the coordinates to the real image. The accuracy rate is 64.0% and loss percentage is 0.074.

\bibliographystyle{unsrt}
\bibliography{main}

\end{document}